\documentclass[runningheads]{llncs}
\usepackage{graphicx}
\usepackage{amsmath}
\usepackage{booktabs}
\usepackage{multirow}
\usepackage{hyperref}
\usepackage{float}
\usepackage{array}
\usepackage{xcolor}
\usepackage{cite} 
\usepackage{caption}

\begin{document}

\title{Trust, Safety, and Accuracy: Assessing LLMs for Routine Maternity Advice.}
\titlerunning{LLMs for Maternal Health Education}
\author{Sai Divya Vissamsetty\inst{1} \and
Anagani Bhanusree\inst{1} \and
Rimjhim\inst{1} \and
K VenkataKrishna Rao\inst{1}}
\authorrunning{Sai Divya et al.}

\institute{National Institute of Technology, Warangal, India\\
\email{vs25csr1p10@student.nitw.ac.in, ab25csr1p11@student.nitw.ac.in, rimjhim@nitw.ac.in, kvkrao@nitw.ac.in}}
\maketitle
\vspace{-0.25em} % Adjust this value as needed

\begin{abstract}
Maternal health information accessibility remains a significant concern in India, where inadequate healthcare infrastructure and socio-cultural hesitations often delay medical consultation. Many women hesitate to discuss pregnancy related concerns with doctors due to stigma or privacy issues, increasingly turning to digital platforms like LLMs for everyday health queries.
This study evaluates three leading LLMs: ChatGPT-4o, Perplexity AI, and GeminiAI for their ability to provide medically reliable, readable, and culturally sensitive responses to seventeen pregnancy related questions, including prevalent myths and misconceptions in the Indian context. Readability of the LLMs responses was assessed using various English readability metrics to ensure that information is presented in simple, clear, and easily understandable language, enabling women especially those from diverse educational backgrounds to comprehend and apply the health guidance effectively. Cosine similarity was used to evaluate semantic alignment with expert medical responses, and Jaccard similarity measured noun-entity overlap. Results show that ChatGPT generated the most readable and clinically coherent responses (FRE = 64.63; FKGL = 7.53), while Perplexity AI achieved the highest semantic similarity (mean cosine = 0.206). For noun overlap, ChatGPT exhibited stronger contextual alignment (Jaccard = 0.172–0.216).
Overall, the findings underscore the potential of LLMs, particularly ChatGPT and Perplexity AI, as supportive tools for addressing pregnancy related myths, enhancing maternal health communication, and improving digital health literacy in underserved Indian communities.

\end{abstract}

\section{Introduction}
\vspace{-2mm}
The issue of maternal health in India persists as a major public health concern, largely due to unequal healthcare access between rural and urban populations, which contributes to delays in essential medical care. Social stigma, limited awareness, and cultural taboos frequently discourage women from discussing reproductive or pregnancy related concerns openly with healthcare providers. At the same time, the increasing reliance on digital platforms for everyday queries ranging from lifestyle advice to health information has led to growing engagement with LLMs. These models are increasingly being used as first line sources of information, even on sensitive topics like pregnancy.
While digital accessibility has expanded rapidly with over 830 million internet users in India and nearly half of rural women now online misinformation and culturally ingrained myths related to pregnancy remain widespread. In such a context, LLMs have the potential to provide quick, empathetic, and privacy preserving responses, but their reliability and cultural sensitivity require systematic evaluation.
Recent studies have explored the use of LLMs in maternal and reproductive healthcare, focusing on reliability, readability, and practical utility.

Lima et al.~\cite{lima2025quality} evaluated LLMs for delivering maternal health information in resource limited settings, highlighting both promise and challenges. Khromchenko et al.~\cite{khromchenko2024chatgpt} compared ChatGPT-3.5 and Gemini on pregnancy questions, noting differences in accuracy and reliability. Onder et al.~\cite{onder2024evaluation} assessed ChatGPT-4’s responses on hypothyroidism during pregnancy, stressing the need for context specific evaluation. Recker et al.~\cite{recker2025large} examined AI integration in gynecological care to aid patient decision making. Insuk et al.~\cite{insuk2025well} showed AI models matching human performance in systematic review screening. Taşkum et al.~\cite{tacskum2025assessment} found high scientific reliability but poor readability in prenatal screening content from LLMs. Wan et al.~\cite{wan2023chatgpt} revealed limitations in accuracy and reference quality of ChatGPT responses. RimJhim et al. \cite{dandapat2022gender} specifies digital platforms, including social media and LLM interactions, have been shown to reflect how cultural norms influence health queries and the spread of misinformation.

Here, we investigated the effectiveness of leading LLMs ChatGPT, Gemini, and Perplexity in addressing pregnancy related queries within the Indian socio-cultural context. These queries covering early gestation to postnatal care, including prevalent myths, were evaluated for accuracy, readability, and contextual understanding against expert medical opinions. Results indicated that LLMs can provide reliable, culturally sensitive, and comprehensible information, supporting maternal health communication in rural and semi-urban regions. The findings emphasize the potential of AI driven tools to complement healthcare services, bridge information gaps, and empower women to make informed decisions through accessible and confidential digital platforms.

\section{Data Collection}

We conducted a structured two-phase methodology to examine pregnancy related information from LLMs and qualified medical practitioners. Seventeen questions including prevalent myths and misconceptions in the Indian context addressing topics from early gestation to postnatal care were designed for comprehensive coverage. In the first phase, responses were obtained by  from leading LLMs to assess their grasp of maternal health.The LLMs were instructed to mimic as a local experienced obstetric and maternal care expert . The second phase involved collecting corresponding answers from experienced obstetric and maternal care experts for comparative analysis. All responses were evaluated using predefined standards for medical accuracy, clarity, and clinical appropriateness. Ethical protocols were maintained throughout the process to ensure participant confidentiality and data reliability. 
\vspace{-1 em}
\section{Methodology}
\vspace{0.1 em}
To assess the quality and reliability of LLM generated responses against those from healthcare professionals, a structured multi-level evaluation framework was implemented. The methodology emphasized core parameters like  readability, semantic similarity and noun entity overlap. 
\vspace{-1 em}
\subsubsection{A) Readability Assessment Using Linguistic Metrics:}
Readability was analyzed to ensure accessibility for non-expert audiences using established linguistic metrics that evaluate sentence complexity, word choice, and syllable count. These measures helped determine the clarity and comprehension level of the generated responses from the LLMs.
The metrics used in this study included:

\textbf{Flesch Reading Ease (FRE):}
This score evaluates how easy a passage is to read, using sentence length and word syllable count. A higher score indicates more accessible language.

\textbf{Flesch-Kincaid Grade Level (FKGL):}
The FKGL is computed to translates text complexity into a U.S. grade level.A score of 8.0 means the content is suitable for someone at the 8\textsuperscript{th}-grade reading level. Lower scores are better for general audiences.

\textbf{Gunning Fog Index (GFI):}
This index estimates the years of formal education required to understand the text. If the score are in the range 7--10 is suitable for general audiences and above 12 college-level or higher. 

\textbf{SMOG Index:}
Often used in public health, the SMOG formula focuses on the number of polysyllabic words. The result reflects the minimum grade level required to understand the text.

\textbf{Dale--Chall Readability Score (DCRS):}
This metric considers how many unfamiliar words appear in a passage. Scores below 5.0 suggest high accessibility.

\textbf{Automated Readability Index (ARI):}
It uses character count to estimate grade level. By applying these metrics, we were able to objectively assess the reading difficulty of each response. 
\vspace{-1 em}
\subsubsection{B) Semantic Similarity Analysis Using Cosine Similarity:}This technique is  used  to compare the meaning of two text samples by analyzing their vector representations. If the scores are close to 1 indicates high similarity in meaning and a score near 0 indicates little or no semantic alignment. 
\vspace{-1 em} % Adjust this value as needed
\subsubsection{C) Noun Entity Recognition Overlap Using Jaccard Similarity:}To quantify how much overlap existed between the identified entities from both sources. A score of 1 means complete overlap (both responses mention the same concepts) and a score of 0 means no shared terminology.
\vspace{-4mm}
\section{Results and Discussion}
\vspace{-2mm}
\subsection{Readability Metrics}
We conducted a thorough analysis of the readability of responses generated by each LLM. This step was crucial to assess how accessible and understandable the responses would be for non-expert readers, particularly expectant mothers and mothers seeking reliable information about pregnancy.
To this end, six widely recognized readability metrics were employed. These metrics can be broadly grouped based on how their scores should be interpreted:

\textbf{Group 1: Metrics Where a Higher Score Indicates Better Readability:}
FRE assesses how easy a text is to read, based on sentence length and syllables per word. A higher FRE score indicates that the content is easier to understand.

From Table.~\ref{tab:readability}, we observe that ChatGPT received the highest FRE score, suggesting that its outputs were the easiest to read among all models. In this metric, a score between 60 and 70 typically indicates that the text is understandable to readers at a middle school (13--15 years old) reading level, which aligns well with public health communication standards.

\textbf{Group 2: Metrics Where a Lower Score Indicates Better Readability:}
This group includes metrics that estimate the minimum grade level required to comprehend the responses from LLMs. A lower score is preferable, as it implies that the response is more accessible to a wider audience.

 From Table.~\ref{tab:readability}, we notice that  ChatGPT produced the most accessible responses, suitable for middle school reading levels. ChatGPT consistently outperformed than Perplexity and Gemini across five readability metrics of Group 2, achieving the lowest scores where simpler text is preferred.
 \vspace{-2 em}
\begin{table}[htbp]
\centering
\setlength{\tabcolsep}{12pt} % Default is 6pt, increase for wider spacing
\renewcommand{\arraystretch}{1.2} % Optional: increases row height slightly
\caption{Average Readability Metrics Across LLMs}
\label{tab:readability}
\begin{tabular}{llccc}
\toprule
\textbf{Group} & \textbf{Metric} & \textbf{ChatGPT} & \textbf{Perplexity} & \textbf{GeminiAI} \\
\midrule
Group 1 & FRE          & \textbf{64.63} & 53.89 & 58.11 \\
\midrule
\multirow{5}{*}{Group 2} 
 & FKGL        & \textbf{7.53}  & 9.67  & 8.24  \\
 & GFI         & \textbf{9.86}  & 12.35 & 10.59 \\
 & SMOG Index  & \textbf{10.04} & 12.00 & 10.83 \\
 & DCRS        & \textbf{9.19}  & 10.10 & 9.47  \\
 & ARI         & \textbf{8.30}  & 10.06 & 8.54  \\
\bottomrule
\end{tabular}
\end{table}
 Perplexity’s responses featured longer sentences and more complex vocabulary, resulting in higher reading levels. Gemini performed moderately but did not exceed ChatGPT. These results emphasize ChatGPT's strength in producing clear, accessible language, which is crucial for sensitive medical topics like pregnancy.
\vspace{-1 em}
\subsection{Semantic Similarity Evaluation Using Cosine Similarity}
To evaluate the semantic alignment between the responses generated by different LLMs and those provided by doctor's responses (D1, D2, D3, D4), we employed cosine similarity as a quantitative metric. It measures how similar two pieces of text are based on their vector representations which capture the semantic content of the text and allow for a comparison that goes beyond simple word matching.
Fig.~\ref{fig:cosine} shows perplexity achieves the highest average similarity in three of the four doctors responses. All three LLMs show a significant drop in similarity with D4, making it the least similar overall. ChatGPT performs slightly better than Perplexity on D2 (0.191 vs. 0.190), though their scores are nearly identical.
\vspace{-1 em}
\subsection{Noun Entity Recognition Overlap Analysis Using Jaccard Similarity}
\begin{figure}[h!]
    \centering
    \begin{minipage}[t]{0.45\textwidth}
        \centering
        \includegraphics[width=\textwidth, height=0.65\textwidth, keepaspectratio]{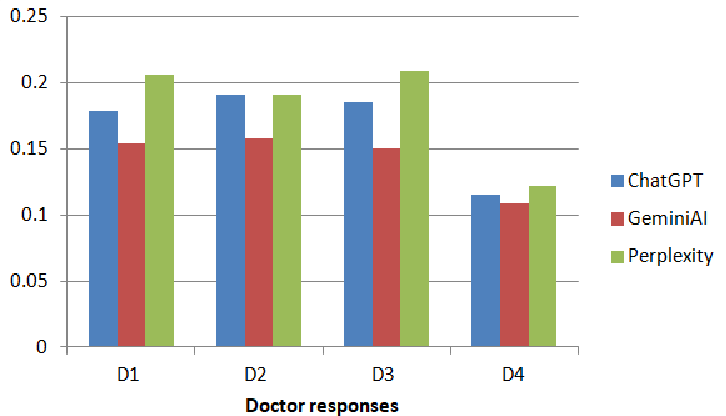}
        \caption{Average Cosine Similarity between LLM and Doctor Responses}
        \label{fig:cosine}
    \end{minipage}%
    \hfill
    \begin{minipage}[t]{0.45\textwidth}
        \centering
        \includegraphics[width=\textwidth, height=0.65\textwidth, keepaspectratio]{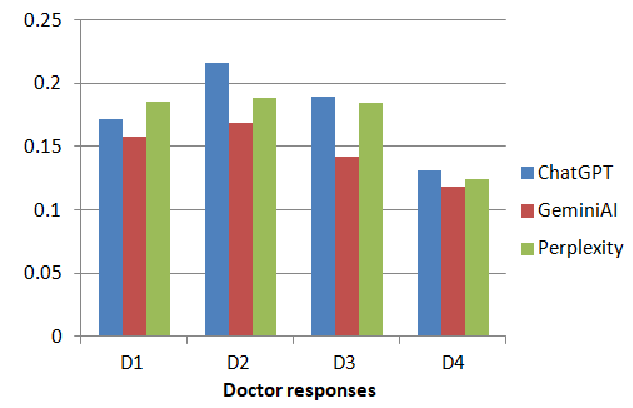}
        \caption{Average Jaccard Similarity between LLM and Doctor Responses}
        \label{fig:jaccard}
    \end{minipage}
\end{figure}

Here, in this method we evaluated how well LLMs captured key noun entities such as medical terms, conditions, anatomical references, and procedures when compared to responses given by medical experts.This metric helps quantify the extent to which an LLM includes the same core clinical concepts as a trained healthcare professional.
Fig.~\ref{fig:jaccard} shows that ChatGPT proved the most effective overall, boosted by a high match on D2, while Perplexity was the most consistently strong model across D1, D2, and D3. All three models struggled to find noun overlap with D4.
\vspace{-3 em}
\section{Conclusion and Future Work}
This study assessed ChatGPT, Perplexity, and Gemini in generating pregnancy related healthcare information using semantic similarity, noun overlap, and readability metrics. ChatGPT performed best, providing accurate, clear, and culturally sensitive responses that effectively addressed common pregnancy myths.
Overall, the responses produced by the LLMs maintained comparatively high readability levels, which may hinder comprehension among women from diverse backgrounds both in rural and semi-urban regions. Future research should expand evaluations to diverse medical domains and socio-cultural contexts for greater inclusivity. Incorporating local health guidelines and domain specific tuning can further enhance accuracy and relevance. Continuous benchmarking and real world validation remain essential for responsible AI use in maternal health communication.
\bibliographystyle{IEEEtran}    
\bibliography{reference} 

@article{lima2025quality,
  title={Quality assessment of large language models’ output in maternal health},
  author={Lima, Henrique A and Trocoli-Couto, Pedro HFS and Moazzam, Zorays and Rocha, Leonardo CD and Pagano, Adriana and Martins, Felipe F and Brabo, Lucas T and Reis, Zilma SN and Keder, Lisa and Begum, Aliya and others},
  journal={Scientific Reports},
  volume={15},
  number={1},
  pages={22474},
  year={2025},
  publisher={Nature Publishing Group UK London}
}

@article{khromchenko2024chatgpt,
  title={ChatGPT-3.5 versus Google Bard: which large language model responds best to commonly asked pregnancy questions?},
  author={Khromchenko, Keren and Shaikh, Sameeha and Singh, Meghana and Vurture, Gregory and Rana, Rima A and Baum, Jonathan D and Rana, Rima},
  journal={Cureus},
  volume={16},
  number={7},
  year={2024},
  publisher={Cureus}
}

@article{onder2024evaluation,
  title={Evaluation of the reliability and readability of ChatGPT-4 responses regarding hypothyroidism during pregnancy},
  author={Onder, CE and Koc, G and Gokbulut, P and Taskaldiran, I and Kuskonmaz, SM},
  journal={Scientific reports},
  volume={14},
  number={1},
  pages={243},
  year={2024},
  publisher={Nature Publishing Group UK London}
}

@article{recker2025large,
  title={Large language models and women’s health: a digital companion for informed decision-making},
  author={Recker, Florian and Neubauer, Ricarda and Wittek, Agnes and Scholten, Nadine},
  journal={Archives of Gynecology and Obstetrics},
  pages={1--8},
  year={2025},
  publisher={Springer}
}

@article{insuk2025well,
  title={How well do ChatGPT and Claude perform in study selection for systematic review in obstetrics},
  author={Insuk, Suppachai and Boonpattharatthiti, Kansak and Booncharoen, Chimbun and Chaipitak, Panitnan and Rashid, Muhammed and Veettil, Sajesh K and Lai, Nai Ming and Chaiyakunapruk, Nathorn and Dhippayom, Teerapon},
  journal={Journal of Medical Systems},
  volume={49},
  number={1},
  pages={1--9},
  year={2025},
  publisher={Springer}
}

@article{tacskum2025assessment,
  title={Assessment of readability, reliability, and quality of large language models in addressing frequently asked questions regarding prenatal screening for fetal chromosomal anomalies},
  author={Ta{\c{s}}kum, {\.I}brahim and S{\i}nac{\i}, Selcan and Aslan, Ferhat and Ta{\c{s}}kum, G{\"u}ls{\"u}m Mina and Sucu, Seyhun},
  journal={International Journal of Gynecology \& Obstetrics},
  year={2025},
  publisher={Wiley Online Library}
}

@article{wan2023chatgpt,
  title={ChatGPT: an evaluation of AI-generated responses to commonly asked pregnancy questions},
  author={Wan, Christopher and Cadiente, Angelo and Khromchenko, Keren and Friedricks, Natalie and Rana, Rima A and Baum, Jonathan D},
  journal={Open Journal of Obstetrics and Gynecology},
  volume={13},
  number={9},
  pages={1528--1546},
  year={2023},
  publisher={Scientific Research Publishing}
}

@article{dandapat2022gender,
  title={Is gender-based violence a confluence of culture? Empirical evidence from social media},
  author={Dandapat, Sourav and others},
  journal={PeerJ Computer Science},
  volume={8},
  pages={e1051},
  year={2022},
  publisher={PeerJ Inc.}
}
\end{document}